\documentclass[letterpaper, 10 pt, conference]{ieeeconf}  

\IEEEoverridecommandlockouts                              

\usepackage{graphicx}
\usepackage{subfig}
\usepackage{amsmath,amssymb} 
\usepackage{color}
\usepackage{tabularx}
\usepackage{amsmath}
\usepackage{cite}
\usepackage{lipsum}
\usepackage{mathtools}
\usepackage{hyperref}
\usepackage[percent]{overpic}

\definecolor{dgreen}{rgb}{0,0,0}
\definecolor{dyellow}{rgb}{.7,.7,0}
\definecolor{dred}{rgb}{1,0,0}
\definecolor{dblue}{rgb}{0,0,0.7}
\definecolor{dorange}{rgb}{0.9,0.5,0.1}

\title{\LARGE \bf
Self-supervised 3D Shape and Viewpoint Estimation \\ from Single Images for Robotics
}

\author{Oier Mees, Maxim Tatarchenko, Thomas Brox and Wolfram Burgard
\thanks{All authors are with the University of Freiburg, Germany. Wolfram Burgard is also with the Toyota Research Institute, Los Altos, USA. This work has partly been supported by the Freiburg Graduate School of Robotics and the Cluster of Excellence BrainLinks-BrainTools under grant EXC-1086.}
}

\begin{document}

\maketitle
\thispagestyle{empty}
\pagestyle{empty}

\begin{abstract}

We present a convolutional neural network for joint 3D shape prediction and viewpoint estimation from a single input image.
During training, our network gets the learning signal from a silhouette of an object in the input image - a form of self-supervision.
It does not require ground truth data for 3D shapes and the viewpoints.
Because it relies on such a weak form of supervision, our approach can easily be applied to real-world data.
We demonstrate that our method produces reasonable qualitative and quantitative results on natural images for both shape estimation and viewpoint prediction.
Unlike previous approaches, our method does not require multiple views of the same object instance in the dataset, which significantly expands the applicability in practical robotics scenarios.
We showcase it by using the hallucinated shapes to improve the performance on the task of grasping real-world objects both in simulation and with a PR2 robot.

\end{abstract}

\section{INTRODUCTION}

The ability to reason about the 3D structure of the world given 2D images only (3D awareness) is an integral part of intelligence that is useful for a multitude of robotics applications.
Making decisions about how to grasp an object~\cite{yan2017learning}, reasoning about object relations~\cite{mees17iros}, anticipating what is behind an object~\cite{varley2017shape} are only a few out of many tasks for which 3D awareness plays a crucial role.
Modern Convolutional Neural Networks (ConvNets) have enabled a rapid boost in single-image 3D shape estimation.
Unlike the classical Structure-from-Motion methods, which infer 3D structure purely based on geometric constraints, ConvNets efficiently build shape priors from data and can rely on those to hallucinate parts of objects invisible in the input image.

Such priors can be learned very efficiently in a fully-supervised manner from 2D image-3D shape pairs~\cite{choy_eccv16, girdhar_eccv16, fan_cvpr17, tatarchenko_iccv17}.
The main limiting factor for exploiting this setup in practical robotics applications is the need for large collections of corresponding 2D images and 3D shapes, which are extremely hard to obtain.
This binds prior methods to training on synthetic datasets, subsequently leading to serious difficulties in their application to real-world tasks.
There are two possible groups of solutions to this problem.
One is to minimize the domain shift, such that models trained using synthetic data could be applied to real images.
The other one suggests exploring weaker forms of supervision, which would allow direct training on real data without going through the dataset collection effort - the direction we pursue in this work. This makes it attractive for many robotic scenarios, for example reasoning about the 3D shapes and poses of objects in a tabletop scene, without the need for ground truth 3D models and the corresponding textures to estimate them at test time nor to train the model.

\begin{figure}
\centering
\includegraphics[width=\linewidth]{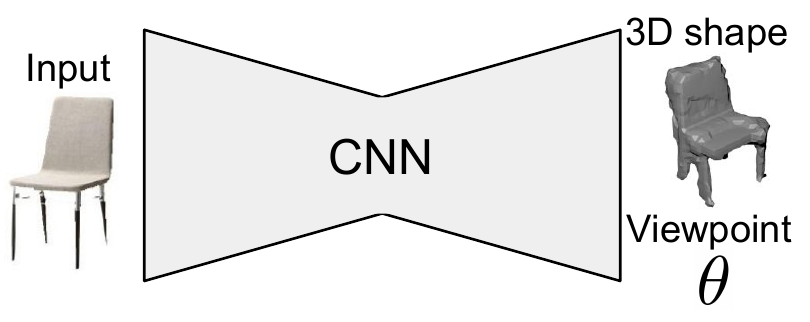}
   \caption{The goal of our work is to predict the viewpoint and the 3D shape of the object from a single image of an object. Our network learns to solve the task solely from matching the segmentation mask of the input object with the projection of the predicted shape.}
\label{fig:teaser}
\end{figure}

Recently, there has been a shift towards learning single-image 3D shape inference using a more natural form of supervision~\cite{yan_nips16, tulsiani_cvpr17, kanazawa_eccv18, tulsiani_eccv18}.
Instead of training on ground truth 3D shapes~\cite{wu20153d}, these methods receive the learning signal from a set of 2D projections of objects.
While this setup substantially relaxes the requirement of having ground-truth 3D reconstructions as supervision, it still depends on being able to capture multiple images of the same object at known~\cite{yan_nips16, tulsiani_cvpr17} or unknown~\cite{tulsiani_eccv18} viewpoints.

In this paper, we push the limits of single-image 3D reconstruction further and present a method which relies on an even weaker form of supervision.
Our ConvNet can infer the 3D shape and the viewpoint of an object from a single image, see Figure~\ref{fig:teaser}.
Importantly, it only learns this from a silhouette of the object in this image.
Though a single silhouette image does not carry any volumetric information, seeing multiple silhouettes of different object instances belonging to the same category allows to infer which 3D shapes could lead to such projections.
Training a network in this setup requires no more than being able to segment foreground objects from the background, which can be done with high confidence using one of the recent off-the-shelve methods~\cite{he_iccv17}.

During training, the network jointly estimates the object's 3D shape and the viewpoint of the input image, projects the predicted shape onto the predicted viewpoint, and compares the resulting silhouette with the silhouette of the input.
Neither the correct 3D shape, nor the viewpoint of the object in the input image is used as supervision.
The only additional source of information we use is the mean shape of the object class which can easily be inferred from synthetic data~\cite{chang_shapenet} and does not limit the method's real-world applicability.

\begin{figure}
\begin{center}
\includegraphics[width=\linewidth]{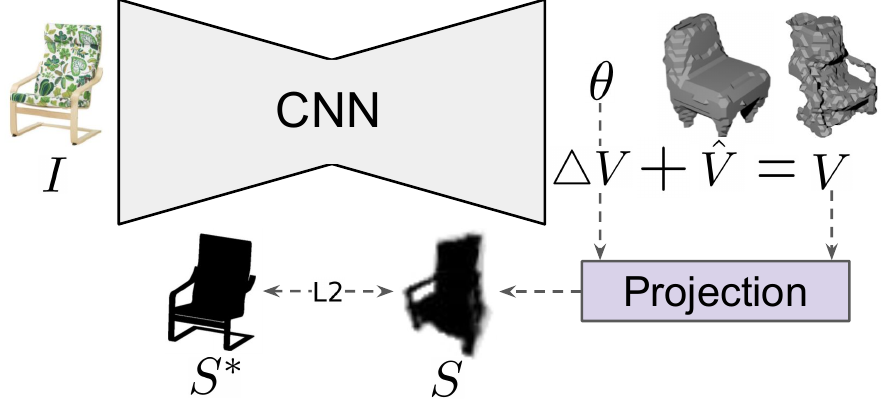}
\end{center}
   \caption{Our encoding-decoding network processes the input RGB image $I$ and predicts the object viewpoint $\theta$ and the shape residual $\triangle V$ which is combined with the mean shape $\hat{V}$ to produce the final estimate $V$ of the 3D model. $V$ is projected onto the predicted viewpoint, the loss between the resulting silhouette image $S$ and the segmentation mask $S^{*}$ of the input image is used to optimize the network.}
\label{fig:architecture}
\end{figure}

We demonstrate both qualitatively and quantitatively that our network trained on synthetic and real-world images successfully predicts 3D shapes of objects belonging to several categories.
Estimated shapes have consistent orientation with respect to a canonical frame.
At the same time, the network robustly estimates the viewpoint of the input image with respect to this canonical frame.
Our approach can be trained with only one view per object instance in the dataset, thus making it readily applicable in a variety of practical robotics scenarios.
We exemplify this by using the reconstructions produced by our method in a robot grasping experiment. Grasp planning based on raw sensory data is difficult due to incomplete scene geometry information.
Relying on the hallucinated 3D shapes instead of the raw depth maps significantly improves the grasping performance of a real-world robot. 


\section{Related work}

Inferring the 3D structure of the world from image data has a long-standing history in computer vision and robotics.
Classical Structure from Motion (SfM) methods were designed to estimate scene geometry from two~\cite{higgins_87} or more~\cite{wu_3dv13} images purely based on geometric constraints.
This class of methods does not exploit semantic information, and thus can only estimate 3D locations for 2D points visible in the input images.
Blanz \emph{et al.}~\cite{blanz_siggraph99} solve the task of 3D reconstruction from a single image using deformable models.
These methods can only deal with a constrained set of object classes and relatively low geometry variations.

More recently, there has emerged a plethora of fully-supervised deep learning methods performing single-image 3D shape estimation for individual objects.
These explore a number of 3D representations which can be generated using ConvNets.
Most commonly, output 3D shapes are represented as voxel grids~\cite{choy_eccv16}.
Using octrees instead of dense voxel grids~\cite{tatarchenko_iccv17, haene_3dv17} allows to generate shapes of higher resolution.
Multiple works concentrated on networks for predicting point clouds~\cite{fan_cvpr17, lin_aaai18} or meshes~\cite{wang_eccv18, groueix_cvpr18}.
In a controlled setting with known viewpoints, 3D shapes can be produced in the form of multi-view depth maps~\cite{lun_3dv17, tatarchenko_eccv16, richter_cvpr18}.

Multiple works attempt solving the task under weaker forms of supervision.
Most commonly, such methods learn from 2D projections of 3D images taken at predefined camera poses.
Those can come as silhouette images~\cite{yan_nips16} as well as richer types of signals like depth or color~\cite{tulsiani_cvpr17, rezende_nips16}.
Gadelha \emph{et al.}~\cite{gadelha_3dv17} train a probabilistic generative model for 3D shapes with a similar type of supervision.
Kanazawa \emph{et al.}~\cite{kanazawa_eccv18} make additional use of mean shapes, keypoint annotations and texture.
Wu \emph{et al.}~\cite{jiajun_nips17} infer intermediate geometric representations which help solve the domain shift problem between synthetic and real data.

Most related to our approach is the work of Tulsiani \emph{et al.}~\cite{tulsiani_eccv18} (mvcSnP).
They perform joint 3D shape estimation and viewpoint prediction while only using 2D segmentations for supervision.
Their approach is based on consistency between different images of the same object, and therefore requires having access to multiple views of the same training instance.
Following a similar setup, our approach relaxes this requirement and can be trained with only one view per instance which simplifies its application in the real world.
The only additional source of information we use is a category-specific mean shape which is easy to get for most of the common object classes.

\section{Method description}

In this section we describe the technical details of our single-image 3D shape estimation and viewpoint prediction method.
The architecture of our system is shown in Figure~\ref{fig:architecture}.

At training time, we encode the input RGB image $I$ into a latent space embedding with an encoder network.
We then process this embedding by two prediction modules: the shape decoder, the output of which is used to reconstruct the final 3D shape $V$, and the viewpoint regressor which predicts the pose $\mathbf{\theta}$ of the object shown in the input image.
The output 3D shape is represented as a $32 \times 32 \times 32$ voxel grid decomposed into the mean shape $\hat{V}$ and the shape residual $\triangle V$.

\begin{equation}
    V = \hat{V} + \triangle V
\label{eq:shape_composition}
\end{equation}

We pre-compute $\hat{V}$ separately for each category and predict $\triangle V$  by the decoder.
The viewpoint $\mathbf{\theta}$ is parametrized with the two angles $[\theta_{az}, \theta_{el}]$, azimuth and elevation,  of the camera rotation around the center of the object. We predict the azimuth angle in the range [0, 360] and the elevation in the range [0, 40] degrees. The 
predicted shape $V$ is projected onto the predicted viewpoint $\mathbf{\theta}$ to generate a silhouette image $S$.
We optimize the squared Euclidean loss $L$ between the predicted silhouette $S$ and the ground truth silhouette $S^{*}$.

\begin{equation}
L = ||S^{*} - S||_2^2
\label{eq:loss}
\end{equation}

Clearly, a single silhouette image does not carry enough information for learning 3D reconstruction.
However, when evaluated across different object instances seen from different viewpoints, $L$ allows to infer which 3D shapes could generate the presented 3D projections.

At test time, the network processes an input image of a previously unseen object and provides estimates for the viewpoint and the 3D shape.

\subsection{Mean shapes}

\begin{figure}
\centering
\def\arraystretch{0}
\begin{tabular}{cccc}
\vspace*{-1mm}
\includegraphics[width=0.18\linewidth]{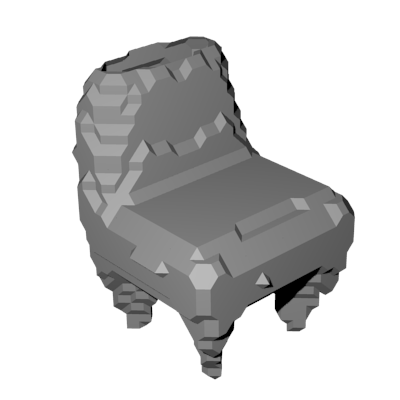}&
\includegraphics[width=0.18\linewidth]{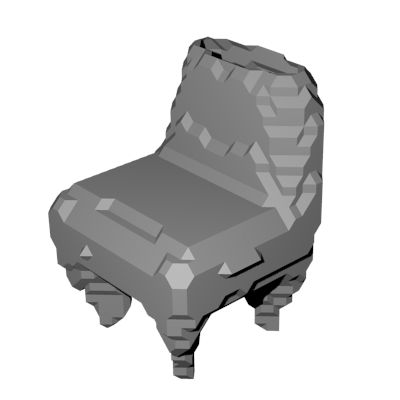}&
\includegraphics[width=0.18\linewidth]{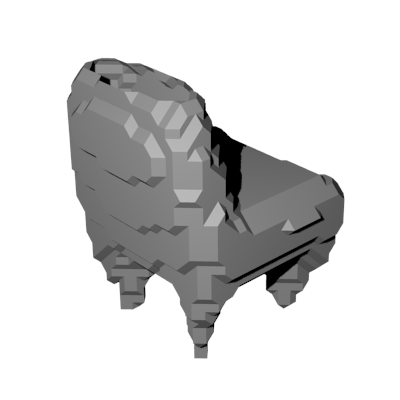}&
\includegraphics[width=0.18\linewidth]{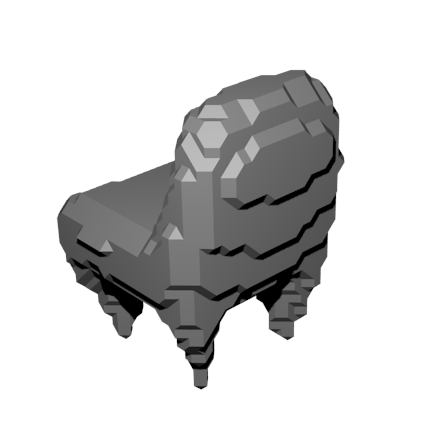}\\
\vspace*{-2mm}
\includegraphics[width=0.18\linewidth]{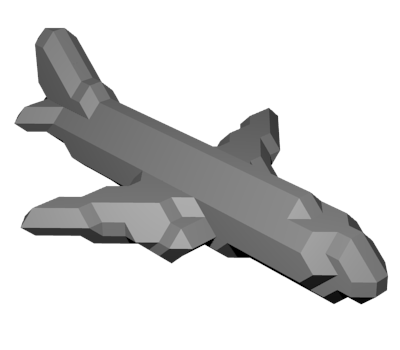}&
\includegraphics[width=0.18\linewidth]{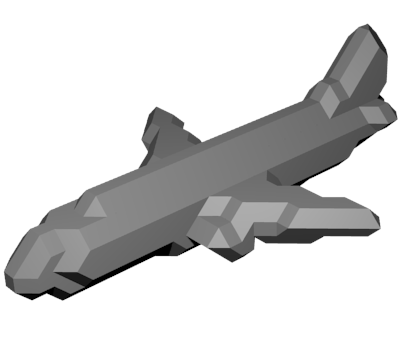}&
\includegraphics[width=0.18\linewidth]{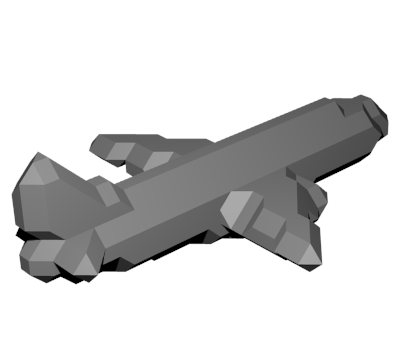}&
\includegraphics[width=0.18\linewidth]{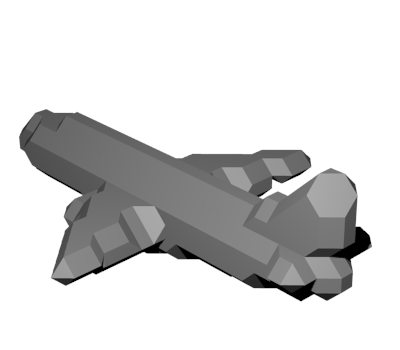}\\
\includegraphics[width=0.18\linewidth]{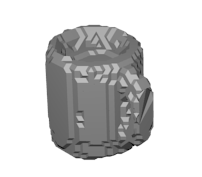}&
\includegraphics[width=0.18\linewidth]{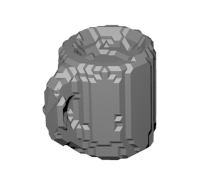}&
\includegraphics[width=0.18\linewidth]{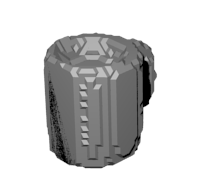}&
\includegraphics[width=0.18\linewidth]{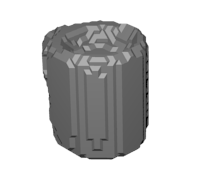}\\
\end{tabular}

\caption{Mean shapes calculated on synthetic data for chairs (top row), planes (middle row) and mugs (bottom row).}
\label{fig:mean_shapes}
\end{figure}

As shown in Equation~\ref{eq:shape_composition}, we represent a 3D shape as a composition of the mean shape $\hat{V}$ and the residual $\triangle V$.
This representation is motivated by two observations.

First, we observed that having a good pose prediction is key to getting reasonable shape reconstructions.
Using the mean shapes allows us to explicitly define the canonical frame of the output 3D shape, which in turn significantly simplifies the job of the pose regressor. One can draw a parallel to Simultaneous Localization and Mapping (SLAM) methods, where on the one hand a good pose estimate makes the mapping easier, and on the other hand an accurate map makes learning the pose estimation better.
In contrast to us, Tulsiani \emph{et al.}~\cite{tulsiani_eccv18} let the network decide what canonical frame to use.
This makes the problem harder and requires carefully designing the optimization procedure for the pose network (using multi-hypotheses output and adversarial training).

Second, it is well-known in the deep learning literature that formulating the task as residual learning often improves the results and makes the training more stable~\cite{he2016deep}.
In our experience, predicting the 3D shape from scratch (without using the mean shape) while also estimating the viewpoint is unstable when only a single view per object instance is available.
Therefore, instead of modeling the full output 3D space with our network, we only model the difference between the mean category shape and the desired shape.  Example mean shapes calculated for three object categories (chairs,  planes and mugs) are shown in Figure~\ref{fig:mean_shapes}.

\begin{figure*}[ht!]
\centering
\vspace*{3mm}
\setlength{\tabcolsep}{1pt}
\begin{tabular}{ccccccccc}
\vspace*{-3.5mm}
\begin{overpic}[width=0.1\linewidth]{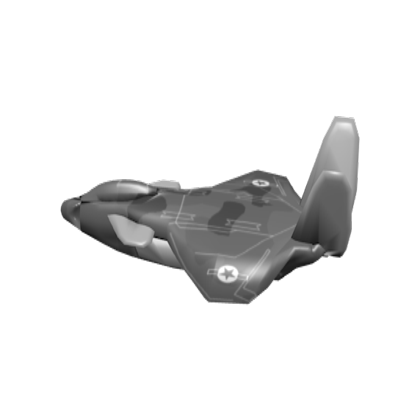}
\put (30,95) {Input}
\end{overpic}&
\begin{overpic}[width=0.1\linewidth]{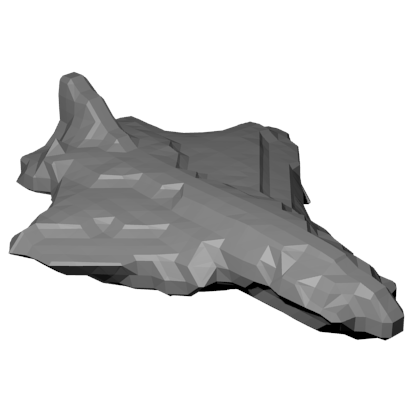}
\put (8,95) {Prediction}
\end{overpic}&
\begin{overpic}[width=0.1\linewidth]{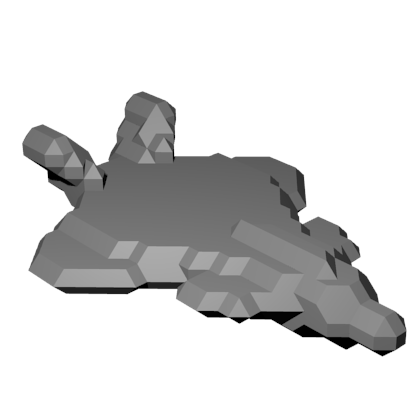}
\put (30,95) {GT}
\end{overpic}&
\begin{overpic}[width=0.1\linewidth]{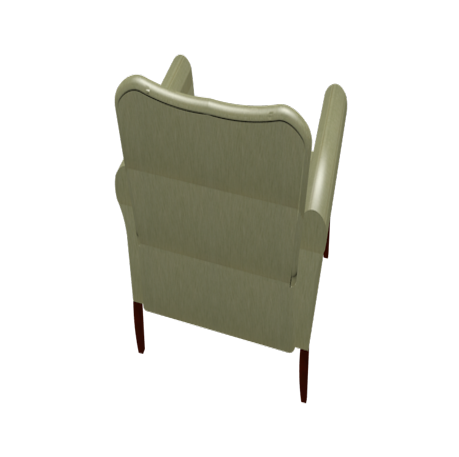}
\put (25,95) {Input}
\end{overpic}&
\begin{overpic}[width=0.1\linewidth]{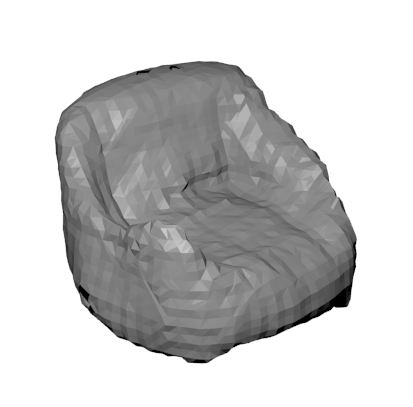}
\put (5,95) {Prediction}
\end{overpic}&
\begin{overpic}[width=0.1\linewidth]{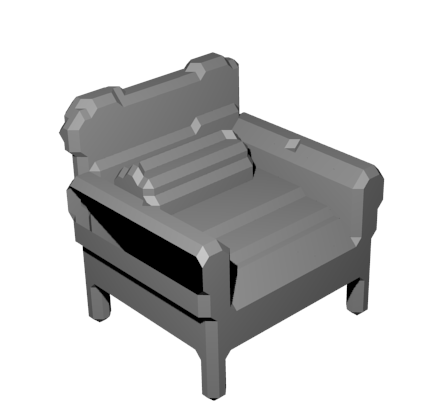}
\put (30,95) {GT}
\end{overpic}&
\begin{overpic}[width=0.1\linewidth]{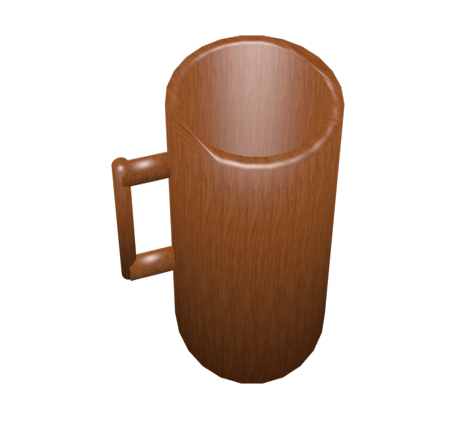}
\put (30,95) {Input}
\end{overpic}&
\begin{overpic}[width=0.1\linewidth]{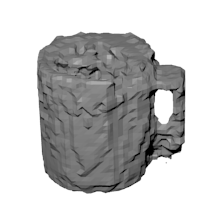}
\put (5,95) {Prediction}
\end{overpic}&
\begin{overpic}[width=0.1\linewidth]{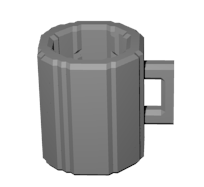}
\put (28,95) {GT}
\end{overpic}\\
\vspace*{-2mm}
\includegraphics[width=0.1\linewidth]{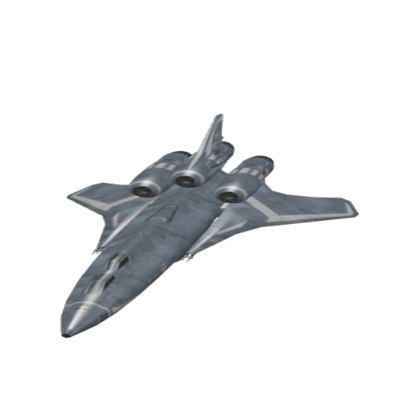}&
\includegraphics[width=0.1\linewidth]{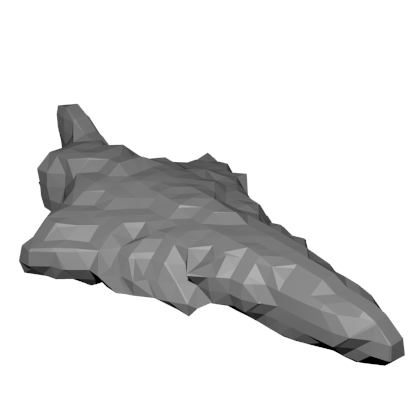}&
\includegraphics[width=0.1\linewidth]{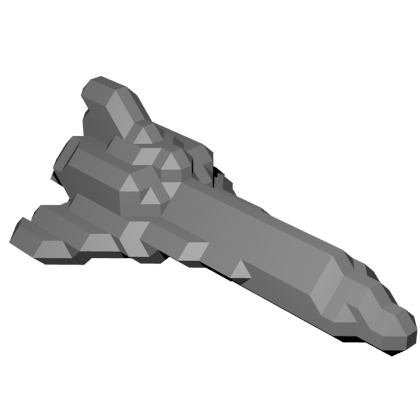}&
\includegraphics[width=0.1\linewidth]{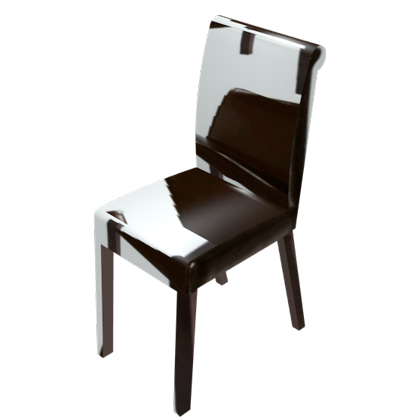}&
\includegraphics[width=0.1\linewidth]{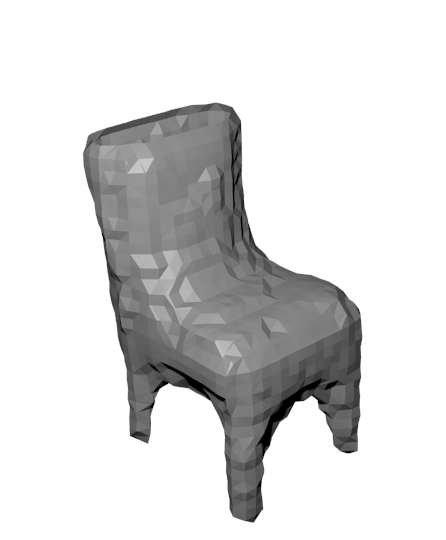}&
\includegraphics[width=0.1\linewidth]{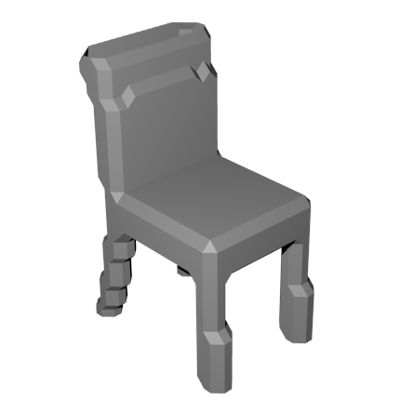}&
\includegraphics[width=0.1\linewidth]{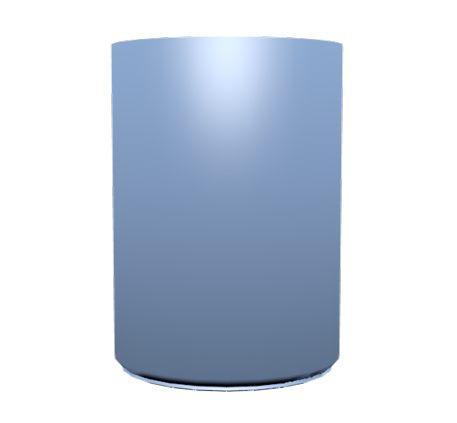}&
\includegraphics[width=0.1\linewidth]{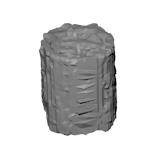}&
\includegraphics[width=0.1\linewidth]{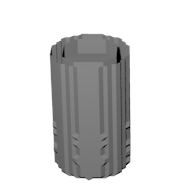}\\
\includegraphics[width=0.1\linewidth]{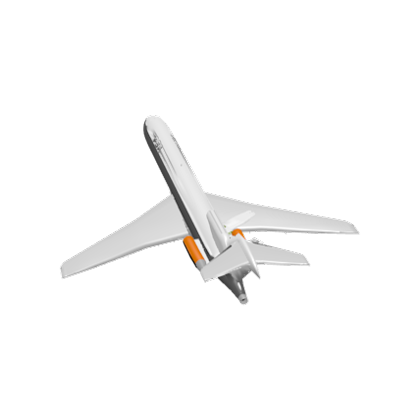}&
\includegraphics[width=0.1\linewidth]{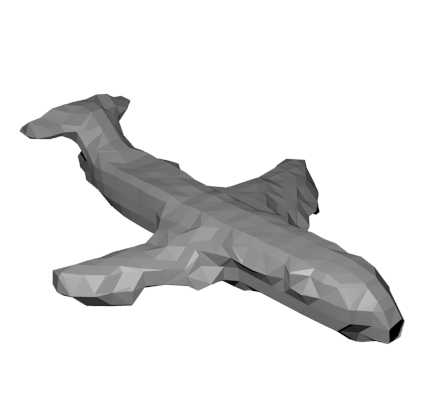}&
\includegraphics[width=0.1\linewidth]{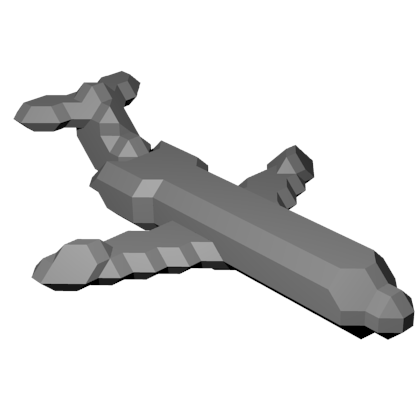}&
\includegraphics[width=0.1\linewidth]{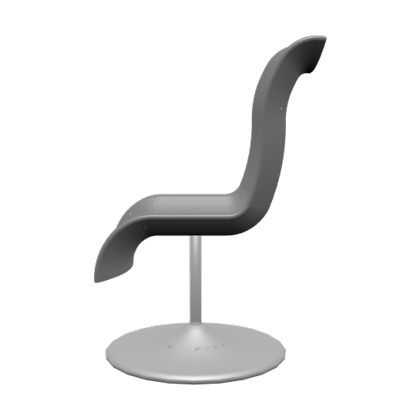}&
\includegraphics[width=0.1\linewidth]{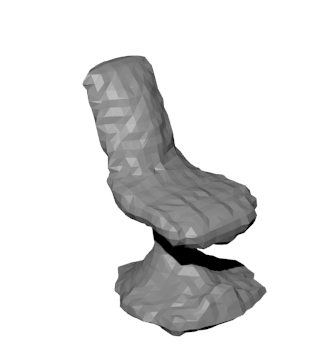}&
\includegraphics[width=0.1\linewidth]{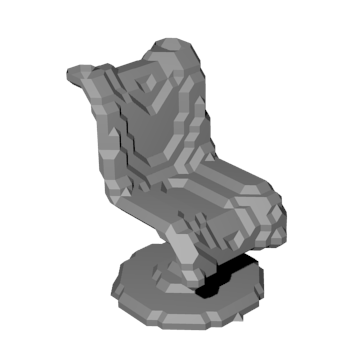}&
\includegraphics[width=0.1\linewidth]{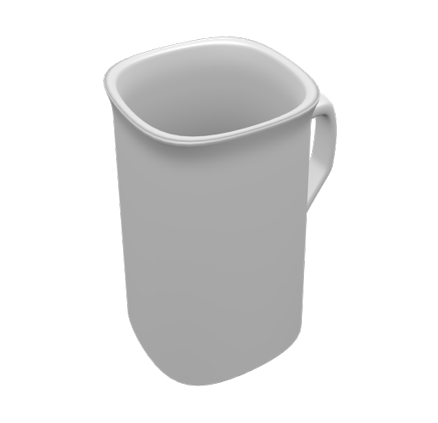}&
\includegraphics[width=0.1\linewidth]{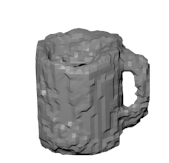}&
\includegraphics[width=0.1\linewidth]{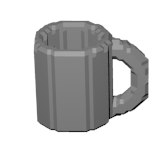}\\
\end{tabular}

\caption{Qualitative results for synthetic images rendered from the Shapenet dataset. Note that we are able to predict 3D shapes that differ substantially from the 3D mean shape.}
\label{fig:synthetic}
\end{figure*}
\subsection{Camera projection}\label{sec:cam}

We generate 2D silhouette images from the predicted 3D shapes using a camera projection matrix $\mathbf{P}$, which can be decomposed into the intrinsic and the extrinsic part.

\begin{equation}
\mathbf{P} = 
\begin{bmatrix}
    \mathbf{K} & \mathbf{0} \\
    \mathbf{0}^T & 1\\
\end{bmatrix}
\begin{bmatrix}
    \mathbf{R} & \mathbf{t} \\
    \mathbf{0}^T & 1\\
\end{bmatrix}
\end{equation}

Following \cite{yan_nips16}, we use a simplified intrinsic camera matrix $\mathbf{K}$ and fix its parameters during training.

\begin{equation}
\mathbf{K} = 
\begin{bmatrix}
    f & 0 & c_x\\
    0 & f & c_y\\
    0 & 0 & 1\\
\end{bmatrix}
\end{equation}

The translation $\mathbf{t}$ represents the distance from the camera center to the object center.
In case of synthetic data, we manually specify $\mathbf{t}$ during rendering.
For real-world images, we crop the objects to their bounding boxes, and re-scale the resulting images to the standard size.
This effectively cancels the translation.

The rotation matrix $\mathbf{R}$ is assembled based on the two angles constituting the predicted viewpoint $\mathbf{\theta}$: elevation $\theta_{el}$ and azimuth $\theta_{az}$.
\begin{equation}
\begin{multlined}
\mathbf{R} = \mathbf{R}_{az} \cdot \mathbf{R}_{el} = \\
\begin{bmatrix}
    \cos\theta_{az} & 0 & \sin\theta_{az}\\
    0 & 1 & 0\\
    -\sin\theta_{az} & 0 & \cos\theta_{az}\\
\end{bmatrix}
\cdot
\begin{bmatrix}
    \cos\theta_{el} & \sin\theta_{el} & 0\\
    -\sin\theta_{el} & \cos\theta_{el} & 0\\
    0 & 0 & 1\\
\end{bmatrix}
\end{multlined}
\end{equation}

Regarding the rendering, similar to spatial transformer networks \cite{jaderberg2015spatial} we perform differentiable volume sampling from the input voxel grid to the output volume and flatten the 3D spatial output along the disparity dimension. Every voxel in the input voxel grid $\mathbf{V} \in \mathbb{R}^{H \times W \times D}$ is represented by a 3D point $(x_{i}^s, y_{i}^s, z_{i}^s)$ with its corresponding occupancy value. Applying the transformation defined by the projection matrix $\mathbf{P}$ to these points generates a set of new locations $(x_{i}^t, y_{i}^t, d_{i}^t)$ in the output volume $\mathbf{U} \in \mathbb{R}^{H' \times W' \times D'}$. We fill out the occupancy values of the output points by interpolating between the occupancy values of the input points:

\begin{multline}
    U_i = \sum_{n}^H\sum_{m}^W\sum_{l}^D V_{nml} \max(0, 1 - |x_{i}^s -m |) \\ \max(0, 1 - |y_{i}^s -n |)\max(0, 1 - |z_{i}^s -l |).\\
\end{multline}

The predicted silhouette $S$ is finally computed by flattening the output volume along the disparity dimension, that is by applying the following $\max$ operator:
\begin{equation}
    S_{n'm'} = \underset{l'}{\max} \  U_{n'm'l'}.
\end{equation}

As we use the $\max$ operator instead of summation, each occupied voxel can only contribute to the foreground pixel of $S$ if it is visible from that specific viewpoint. Moreover, empty voxels will not contribute to the projected silhouette from any viewpoint.

\subsection{Training schedule}

We train our network in multiple stages.
In the first stage, we pre-train the network from scratch on synthetic data.
For the initial 300K iterations we freeze the shape decoder and only train the pose regressor.
This pushes the network to produce adequate pose predictions which happens to be crucial for obtaining reasonable 3D reconstructions.
Not training the pose regressor first results in unstable training, as it is hard for the network to decide how to balance learning the shape reconstruction and the pose regressor jointly from scratch.
After that, we train the entire network, including the shape estimation decoder.

At the second stage, we fine-tune the entire network on real images.
We found that even though the appearance of synthetic and real images is significantly different, pre-training on synthetic data simplifies training on the real-world dataset.
\begin{figure*}[h]
\centering
\vspace*{4mm}
\begin{tabular}{ccccccccc}
\vspace*{-2mm}
\begin{overpic}[width=0.047\linewidth]{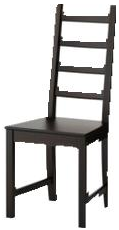}
\put (15,117) {Input}
\end{overpic}&
\begin{overpic}[width=0.1\linewidth]{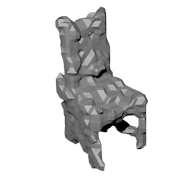}
\put (10,105) {Prediction}
\end{overpic}&
\begin{overpic}[width=0.1\linewidth]{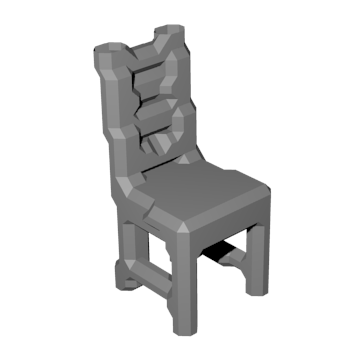}
\put (0,105) {Ground truth}
\end{overpic}&
\begin{overpic}[width=0.065\linewidth]{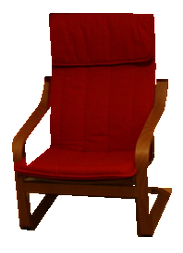}
\put (15,117) {Input}
\end{overpic}&
\begin{overpic}[width=0.1\linewidth]{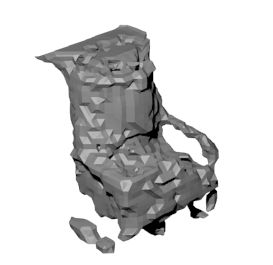}
\put (10,105) {Prediction}
\end{overpic}&
\begin{overpic}[width=0.1\linewidth]{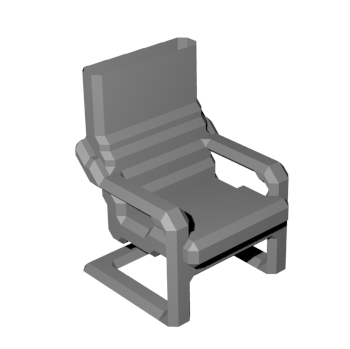}
\put (0,105) {Ground truth}
\end{overpic}&
\begin{overpic}[width=0.06\linewidth]{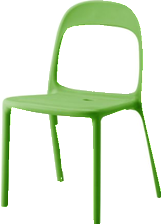}
\put (17,125) {Input}
\end{overpic}&
\begin{overpic}[width=0.1\linewidth]{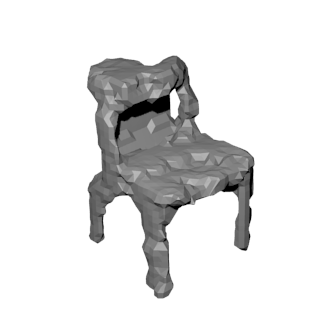}
\put (8,101) {Prediction}
\end{overpic}&
\begin{overpic}[width=0.1\linewidth]{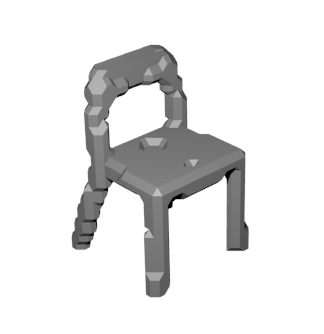}
\put (0,101) {Ground truth}
\end{overpic}\\
\includegraphics[width=0.052\linewidth]{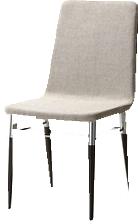}&
\includegraphics[width=0.1\linewidth]{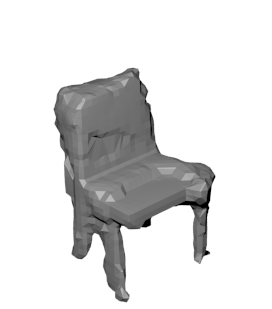}&
\includegraphics[width=0.1\linewidth]{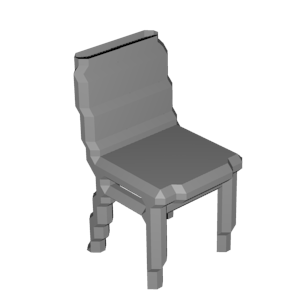}&
\includegraphics[width=0.07\linewidth]{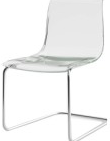}&
\includegraphics[width=0.1\linewidth]{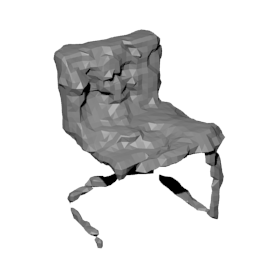}&
\includegraphics[width=0.1\linewidth]{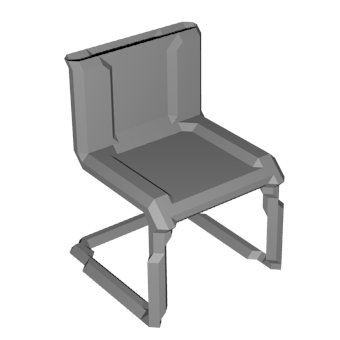}&
\includegraphics[width=0.072\linewidth]{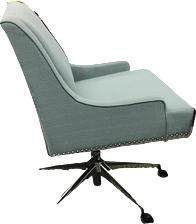}&
\includegraphics[width=0.1\linewidth]{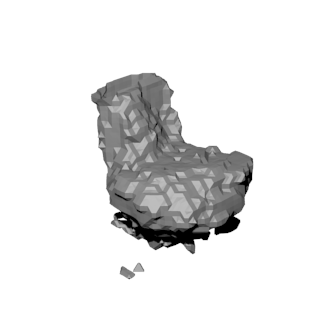}&
\includegraphics[width=0.1\linewidth]{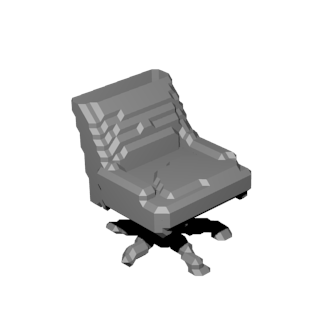}
\end{tabular}

\caption{Qualitative analysis of the predicted shapes on real images. Despite the large variance from synthetic to real data, we are able to successfully predict consistent shapes.}
\label{fig:real}
\end{figure*}

\subsection{Network architecture}

Our network has three components: a 2D convolutional encoder, a viewpoint estimation decoder, and a 3D up-convolutional decoder that predicts the occupancies of the residual in a voxel grid of $32 \times 32 \times 32$. The encoder consists of 3 convolutional layers with 64, 128 and 256 channels. The bottleneck of the network contains 3 fully connected layers of size 512, 512 and 256. The last layer of the bottleneck is fed to the viewpoint estimation block and to the 3D up-convolutional decoder. For the viewpoint estimation we use 2 fully-connected layers to regress the azimuth and the elevation angle. The 3D decoder consists of one fully-connected layer of size 512 and 3 convolutional layers with channel size 256, 128, 1. 

\section{Experiments}

In this section we showcase our approach both qualitatively and quantitatively, and demonstrate its applicability in a real-world setting.

\subsection{Dataset}

We evaluate our approach on both synthetic and real data. For experiments on synthetic data, we use the ShapeNet dataset \cite{chang_shapenet}. It contains around 51,300 3D models from 55 object classes. We pick three representative classes: chairs, planes and mugs. The images are rendered together with their segmentation masks: the azimuth angles $\theta_{az}$ are sampled regularly with a 15 degree step, and the elevation angles $\theta_{el}$ are sampled randomly in the range [0, 40] degrees. We then crop and rescale the centering region of each image to $64 \times 64 \times 3$ pixels. The 3D ground truth shapes are downsampled to a $32 \times 32 \times 32$ voxel grid and oriented to a canonical view.

For experiments on real data, we leverage the chair class from the Pix3D dataset \cite{sun2018pix3d}, which has around 3800 images of chairs. However, many chairs are fully/partially occluded or truncated. We remove those images, as well as images that have an in-plane rotation of more than 10 degrees. For the mugs class we record a dataset of 648 images and point clouds with a RGB-D camera, with similar poses sampled as for ShapeNet. We additionally compute 3D models of the objects by merging together multiple views via Iterative Closest Point. We crop the object images to their  bounding boxes, and re-scale the resulting images to the standard size. This effectively cancels the translation in our projection matrix $\mathbf{P}$, as mentioned in Sec \ref{sec:cam}.

\subsection{Evaluation protocol}

For the final quantitative evaluation of the shape prediction, we report the mean intersection over union (IoU) between the ground truth and the predictions. We binarize the predictions by determining the thresholds from the data, similar to \cite{tulsiani_eccv18}. To stay comparable with \cite{tulsiani_eccv18}, we use two random views per object instance for evaluation.
Viewpoint estimation is evaluated by measuring the angular distance between the predicted and the ground-truth rotation in degrees. Following \cite{tulsiani_eccv18}, we report two metrics: Median Angular Error (Med-Err) and fraction of instances with an error less than 30 degrees (Acc$_{\frac{\pi}{6}}$). To evaluate the similarity of point clouds, we use the Hausdorff distance. We compute the symmetric Hausdorff distance by running it both ways and averaging the distance from the predicted or raw point cloud to its closest point in the ground truth point cloud.

\subsection{Synthetic data}

We started off by evaluating our method on the ShapeNet dataset \cite{chang_shapenet}.
Quantitative results of shape reconstruction are reported in Table \ref{tab:shape}. We compare our method with several baselines that rely on multiple silhouette images of an object instance during training, and also against a baseline model which is trained with full 3D supervision.

\begin{table}[h]
  \centering
  \begin{tabular}{l c c c c}
  Method & Training Views & Planes & Chairs & Mugs\\
  \hline
  \hline
  Ours (3D supervision) & - & 0.57 & 0.53 & 0.45\\
  PTN\cite{yan_nips16} & 24  & x & 0.506  & x\\
  DRC \cite{tulsiani_cvpr17} & 5 & 0.5 & 0.43 & x\\
  mvcSnP\cite{tulsiani_eccv18} & 2 & 0.52 & 0.40 & x\\
  Ours & 1 & 0.47 & 0.36 & 0.40\\
  \end{tabular}
  \caption{Quantitative comparison of multiple single-view 3D reconstruction approaches. Our method yields competitive results though relying on a weaker form of supervision.}
  \label{tab:shape}
\end{table} 

We observe that the performance of the multi-view methods increases as more views are used for training. Having multiple silhouette images allows to better assess the quality of the predicted shape, which constrains the optimization procedure and yields better result. For instance, PTN\cite{yan_nips16} uses 24 views from an object instance to correct the predicted voxel grid in each backward pass. Despite using a single silhouette during training, we obtain the mean IoU scores of 0.47 for planes and 0.36 for chairs. This result is very close to the baselines method that rely on stronger supervision, like  mvcSnP \cite{tulsiani_eccv18} and DRC \cite{tulsiani_cvpr17}.
\begin{table}[h]
  \centering
  \begin{tabular}{l ccc ccc ccc}
  Method   &  \multicolumn{2}{c}{Plane} & \multicolumn{2}{c}{Chair} & \multicolumn{2}{c}{Mug}\\
    \hline
  \hline
   & Err &  Acc$_{\frac{\pi}{6}}$ & Err &  Acc$_{\frac{\pi}{6}}$ & Err &  Acc$_{\frac{\pi}{6}}$\\
\hline
  mvcSnP\cite{tulsiani_eccv18} & 14.3  & 0.69 & 7.8  & 0.81 & x & x \\
 Ours & 7.0  & 0.83 & 10.7 & 0.78  & 14.6 & 0.7\\
  \end{tabular}
  \caption{Analysis of the performance of the viewpoint estimation on the Shapenet dataset.}
  \label{tab:pose}
\end{table}   

We show qualitative examples of the predicted 3D shapes in Figure~\ref{fig:synthetic}. For visualization, voxel grids were converted to meshes using the marching cubes algorithm. Note how we are able to predict shapes which are substantially different from the mean shape.

\begin{figure}[h]
\begin{center}
\includegraphics[width=0.87\linewidth]{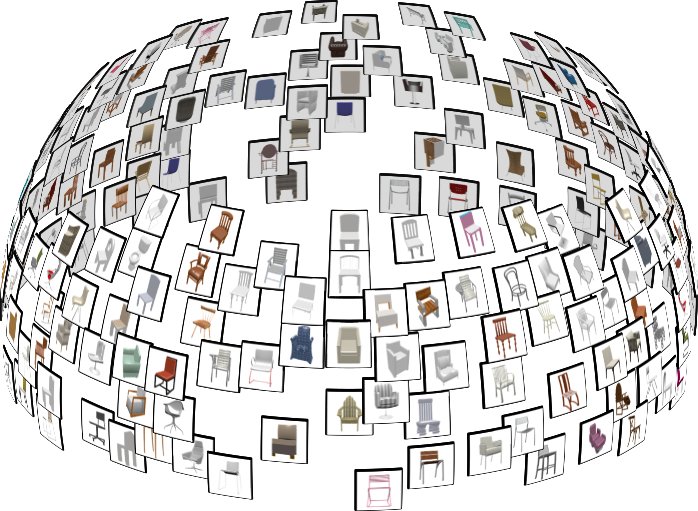}
\end{center}
   \caption{Visualization of the learned pose distribution.}
\label{fig:pose_dist}
\end{figure}

The results of pose prediction are shown in Table~\ref{tab:pose}. For planes, compared to mvcSnP~\cite{tulsiani_eccv18} we observe a relative improvement of 7.3 degrees for the median angle error and a relative improvement of 0.14\% for the fraction of instances with an angular error less than 30 degrees. For chairs we report 10.7 median angular error and additionally we visualize the learned pose distribution in Figure  \ref{fig:pose_dist}. 

\begin{figure}[t]
\centering
\vspace*{4mm}
\setlength{\tabcolsep}{5pt}
\begin{tabular}{cccccc}
\vspace*{-2mm}
\begin{overpic}[width=0.085\linewidth]{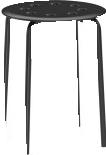}
\put (0,122) {Input}
\end{overpic}&
\begin{overpic}[width=0.15\linewidth]{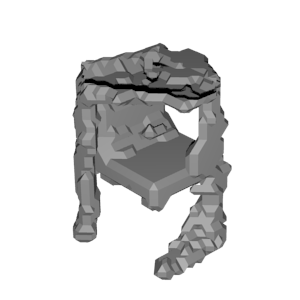}
\put (10,99) {Prediction}
\end{overpic}&
\begin{overpic}[width=0.15\linewidth]{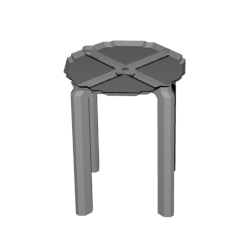}
\put (30,99) {GT}
\end{overpic}&
\begin{overpic}[width=0.085\linewidth]{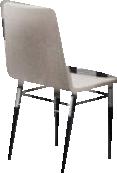}
\put (0,122) {Input}
\end{overpic}&
\begin{overpic}[width=0.15\linewidth]{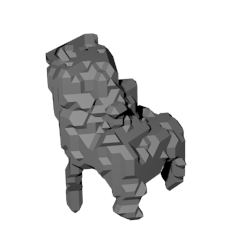}
\put (10,99) {View1}
\end{overpic}&
\begin{overpic}[width=0.15\linewidth]{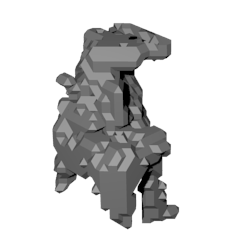}
\put (10,99) {View2}
\end{overpic} \\
\includegraphics[width=0.081\linewidth]{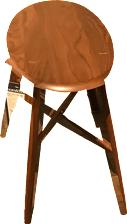}&
\includegraphics[width=0.15\linewidth]{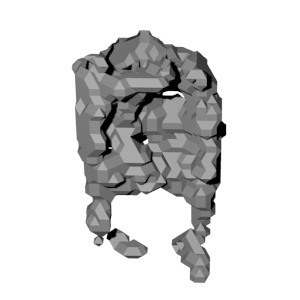}&
\includegraphics[width=0.15\linewidth]{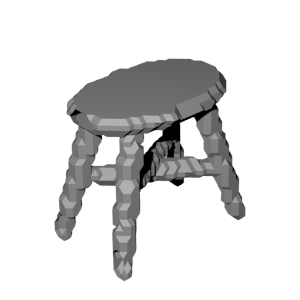}&
\includegraphics[width=0.08\linewidth]{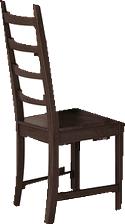}&
\includegraphics[width=0.15\linewidth]{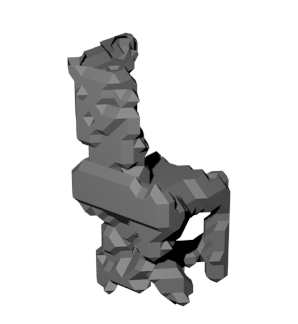}&
\includegraphics[width=0.14\linewidth]{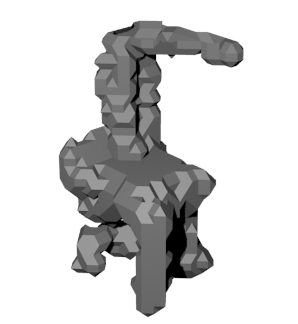}
\end{tabular}

\caption{Certain failure cases. On the top left example the network predicts an approximate shape of the round chair, but fails to remove the mean shape. For the bottom left case, the network fails to predict a consistent shape. On the right, it reconstructs shapes which only fit the input view and look incorrect from unobserved viewpoints.}
\label{fig:failure}
\end{figure}
\subsection{Real data}

We also evaluate our approach on real-world images. 
We learn to predict the shapes and poses of the real-world chairs by fine-tuning our model trained on synthetic images. Despite the large domain gap between synthetic and real images, we achieve a mean IoU of 0.21 and a median angular pose error of 0.27 in this challenging setting. We show qualitative results in Figure~\ref{fig:real}. In this extremely challenging setting, the network learns to produce non-trivial reconstructions that differ significantly from the mean shape.

We also show failure cases in Figure~\ref{fig:failure}. In the top left row the network predicts an approximate shape of the round chair but fails to remove the mean shape from it. For the bottom left row, the network is not able to predict a consistent shape. On the right, we show examples where the predicted shape is consistent from the input image view, but looks incorrect from other viewpoints. This is a inherent problem of multi-view 3D reconstruction methods when the number of observations is low.

Overall, these results on the data derived from a challenging real-world setting concretely demonstrate the ability of our approach to learn joint 3D shape and viewpoint estimation despite the absence of direct shape or pose supervision during training. Even though some reconstructions look noisy and lack fine details, due to the inherent shape ambiguity of multi-view based 3D reconstruction approaches when the number of observed views is low, our results are  promising given the extreme complexity of the task.

\subsection{Grasping}

Following the previous setup, we also evaluate our approach on real mugs recorded with a RGB-D camera. Grasp planning based on raw sensory data is difficult due to incomplete scene geometry information. We leverage the ability of our approach to hallucinate the object parts, such as the mug handles, that are not visible, to improve grasping performance.

First we convert the predicted voxel grids to  point clouds and scale them to match the real world size of the mug. The density of the point cloud is compared to the densities of the real clouds to match it accordingly. The densities are
computed by randomly sampling $\frac{1}{10}$ of the points and averaging
the distances to their nearest neighbors. The raw partial point clouds have a Hausdorff distance of 8.7 millimeters with respect to the full ground truth mugs, while our predicted mugs have a distance of 3.8 millimeters. 

\begin{figure}[t]

\subfloat[][]{\includegraphics[width=0.3\linewidth]{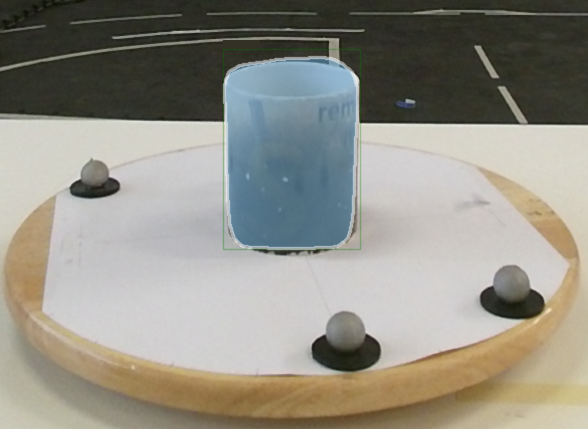}
}
\subfloat[][]{\includegraphics[width=0.3\linewidth]{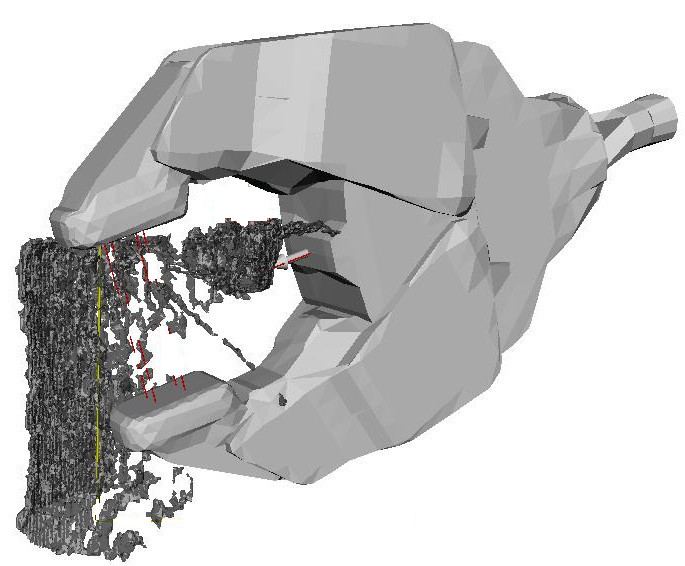}\label{fig:graspb}
}    
\subfloat[][]{\includegraphics[width=0.3\linewidth]{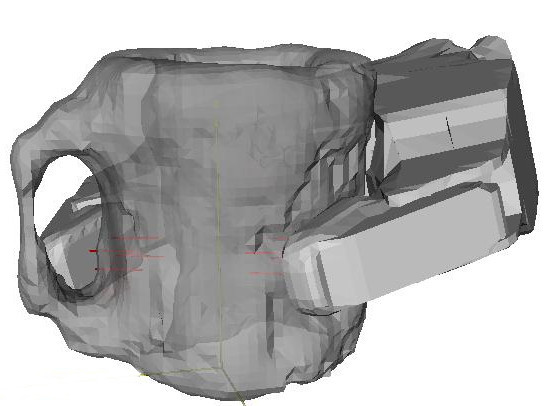}
} 
\caption{Input scene is segmented with Mask-RCNN~\cite{he_iccv17} (a). A grasp planned on the raw sensory data that penetrates the real object (b). Grasp computed on the predicted object (c).} \label{fig:grasp}
\end{figure}

To evaluate grasping performance we leverage the GraspIt!~\cite{miller2004graspit} simulator.  We compute grasps on the meshes of the segmented raw point cloud and on our predictions and choose the highest scoring ones. In order to simulate a real-world grasp execution, the object is removed and replaced with the ground truth mesh. The hand is then placed 15cm away from the ground truth object along the approach direction of the grasp planned in the previous step. The hand is moved along the approach direction of the planned grasp until reaching the grasp pose
or making contact. This helps us determine if the grasp would have been a failure because the grasp penetrates the real object, as seen in Figure \ref{fig:graspb}. We report a grasp success of 63.2\% on the partial clouds and of 82.1\% on the predicted models.
\begin{figure}[b]
\centering
\setlength{\tabcolsep}{0.5pt}
\subfloat[][]{\includegraphics[width=0.32\linewidth]{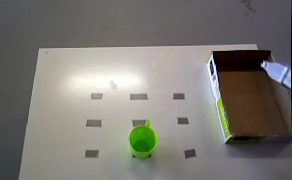}\label{fig:pr2graspa}
}
\subfloat[][]{\includegraphics[width=0.32\linewidth]{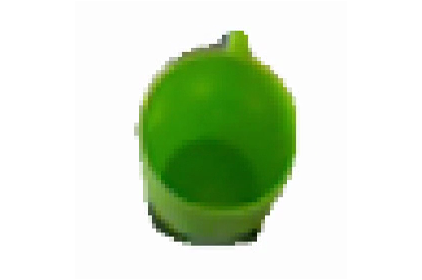}
}    
\subfloat[][]{\includegraphics[width=0.32\linewidth]{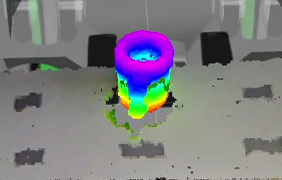}
} 

\begin{tabular}{ccc}
\includegraphics[width=0.33\linewidth]{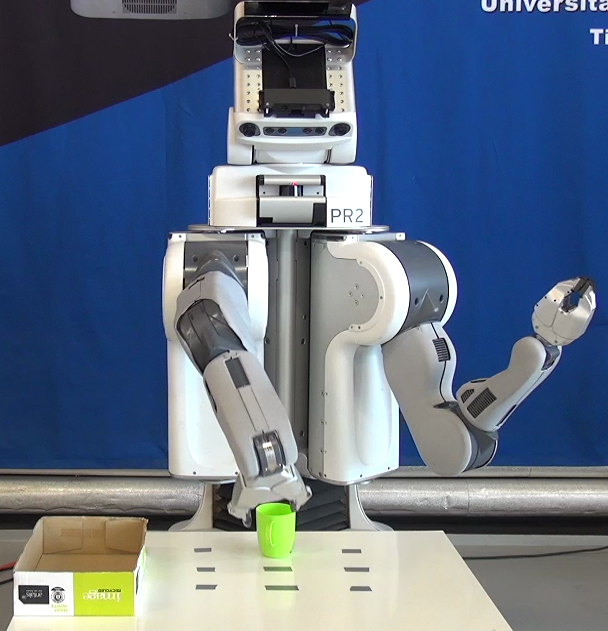}&
\includegraphics[width=0.33\linewidth]{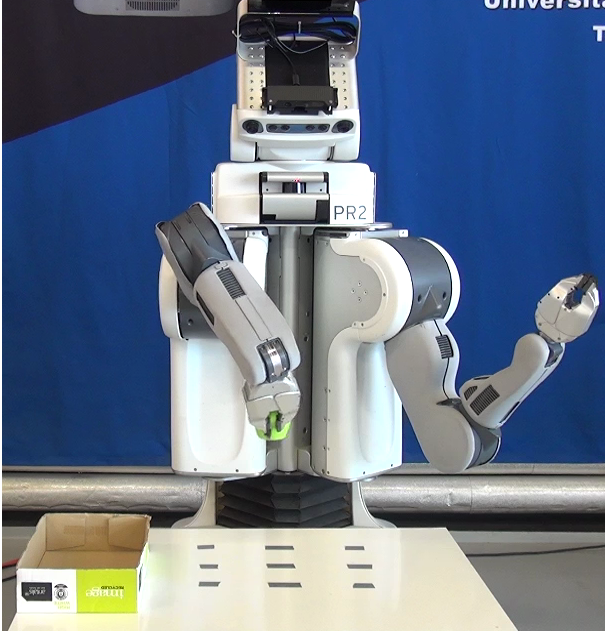}&
\includegraphics[width=0.33\linewidth]{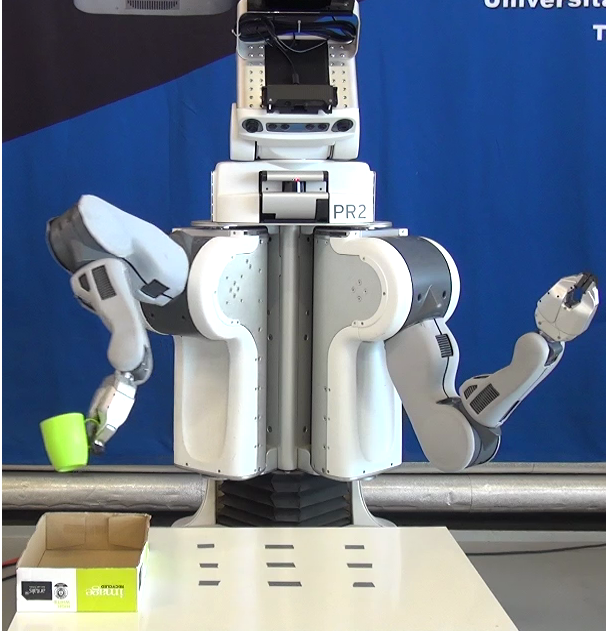}
\end{tabular}
\caption{Example pick-and-place execution with the PR2 robot. We use the  robot's top-down  camera view (a). The scene is segmented with Mask-RCNN~\cite{he_iccv17} to create the input for our network (b). The predicted point cloud (c). The bottom row shows a successful pick-and-place execution based on the predicted point cloud, where the robot grasps the mug by the handle.} \label{fig:pr2grasp}
\end{figure}

To exemplify the ability of our approach to improve grasping performance we use a PR2 robot to perform grasps planned by using Grasp Pose Detection (GPD)~\cite{gualtieri2016high}, which predicts a series of 6-DOF candidate grasp poses given a 3D point cloud for a 2-finger grasp. The reachability of the proposed candidate grasps are checked using MoveIt!~\cite{chitta2012moveit}, and the highest quality reachable grasp is then executed with the PR2 robot. After picking one of the two unseen mugs, we command the robot to place the mug in a box. We mark 9 positions on the table  from which to pick the mugs in various orientations, as shown in Figure \ref{fig:pr2grasp}. We evaluate the success rate by counting the times the robot successfully picked and placed the mug inside the box. We perform 117 grasps per method and report a success rate of 44.4\% on the raw clouds and of 70.9\% on the predicted models. We show an example run in Figure \ref{fig:pr2grasp}. The top row shows the robot's camera view, the segmented image we input to our network and the predicted point cloud. The bottom row shows the PR2 robot successfully grasping the mug by the handle, which was not visible in the input view. Compared to using raw sensor data, our method enables a PR2 robot to improve pick-and-place performance by enabling more precise grasping, such as grasping  mugs by the handle. 



\section{Conclusions and discussion}
In this paper, we presented a novel self-supervised approach to the problem of learning joint 3D shape and pose from a single input image which can be trained with as little as one view per object instance.
We exemplified how the reconstructions produced by our method improve the grasping performance of a real-world robot.

We assumed that every object category can be decently modeled with a single mean shape which of course is not always true.
Addressing this issue would require calculating the mean shapes over sub-categories and combining our category-specific networks with a fine-grained classifier.
Another improvement would be to combine category-specific networks into a single universal reconstruction-and-viewpoint-estimation network.
It is also important to extend the set of predicted poses from two angles to a full 3D rotation. We also note that due to the weak form of supervision used, our approach is exposed to the inherent shape ambiguity of multi-view-based 3D reconstruction approaches when the number of observed views is low. Moreover, training a viewpoint estimator for symmetric objects such as tables and cars is sometimes unstable. Adding a photometric loss or learning view priors that aid constraining the optimization procedure might help alleviating these problems.

\section*{Acknowledgments}
We would like to thank Andreas Wachaja for support while recording the mugs dataset. We thank Andreas Eitel and Nico Hauff for feedback on the grasping experiments.


\bibliographystyle{unsrt}
\bibliography{root}

\begin{thebibliography}{10}

\bibitem{yan2017learning}
Xinchen Yan, Jasmined Hsu, Mohammad Khansari, Yunfei Bai, Arkanath Pathak,
  Abhinav Gupta, James Davidson, and Honglak Lee.
\newblock Learning 6-dof grasping interaction via deep geometry-aware 3d
  representations.
\newblock In {\em IEEE International Conference on Robotics and Automation
  (ICRA)}, 2018.

\bibitem{mees17iros}
Oier Mees, Nichola Abdo, Mladen Mazuran, and Wolfram Burgard.
\newblock Metric learning for generalizing spatial relations to new objects.
\newblock In {\em Proceedings of the International Conference on Intelligent
  Robots and Systems (IROS)}, Vancouver, Canada, 2017.

\bibitem{varley2017shape}
Jacob Varley, Chad DeChant, Adam Richardson, Joaqu{\'\i}n Ruales, and Peter
  Allen.
\newblock Shape completion enabled robotic grasping.
\newblock In {\em IROS}, 2017.

\bibitem{choy_eccv16}
Christopher~Bongsoo Choy, Danfei Xu, JunYoung Gwak, Kevin Chen, and Silvio
  Savarese.
\newblock 3d-r2n2: {A} unified approach for single and multi-view 3d object
  reconstruction.
\newblock In {\em ECCV}, 2016.

\bibitem{girdhar_eccv16}
Rohit Girdhar, David~F. Fouhey, Mikel Rodriguez, and Abhinav Gupta.
\newblock Learning a predictable and generative vector representation for
  objects.
\newblock In {\em ECCV}, 2016.

\bibitem{fan_cvpr17}
Haoqiang Fan, Hao Su, and Leonidas~J. Guibas.
\newblock A point set generation network for 3d object reconstruction from a
  single image.
\newblock In {\em CVPR}, 2017.

\bibitem{tatarchenko_iccv17}
Maxim Tatarchenko, Alexey Dosovitskiy, and Thomas Brox.
\newblock Octree generating networks: Efficient convolutional architectures for
  high-resolution 3d outputs.
\newblock In {\em ICCV}, 2017.

\bibitem{yan_nips16}
Xinchen Yan, Jimei Yang, Ersin Yumer, Yijie Guo, and Honglak Lee.
\newblock Perspective transformer nets: Learning single-view 3d object
  reconstruction without 3d supervision.
\newblock In {\em NIPS}, 2016.

\bibitem{tulsiani_cvpr17}
Shubham Tulsiani, Tinghui Zhou, Alexei~A. Efros, and Jitendra Malik.
\newblock Multi-view supervision for single-view reconstruction via
  differentiable ray consistency.
\newblock In {\em CVPR}, 2017.

\bibitem{kanazawa_eccv18}
Angjoo Kanazawa, Shubham Tulsiani, Alexei~A. Efros, and Jitendra Malik.
\newblock Learning category-specific mesh reconstruction from image
  collections.
\newblock In {\em ECCV}, 2018.

\bibitem{tulsiani_eccv18}
Shubham Tulsiani, Alexei~A. Efros, and Jitendra Malik.
\newblock Multi-view consistency as supervisory signal for learning shape and
  pose prediction.
\newblock In {\em CVPR}, 2018.

\bibitem{wu20153d}
Zhirong Wu, Shuran Song, Aditya Khosla, Fisher Yu, Linguang Zhang, Xiaoou Tang,
  and Jianxiong Xiao.
\newblock 3d shapenets: A deep representation for volumetric shapes.
\newblock In {\em CVPR}, 2015.

\bibitem{he_iccv17}
Kaiming He, Georgia Gkioxari, Piotr Doll{\'{a}}r, and Ross~B. Girshick.
\newblock Mask {R-CNN}.
\newblock In {\em ICCV}, 2017.

\bibitem{chang_shapenet}
Angel~X. Chang, Thomas Funkhouser, Leonidas Guibas, Pat Hanrahan, Qixing Huang,
  Zimo Li, Silvio Savarese, Manolis Savva, Shuran Song, Hao Su, Jianxiong Xiao,
  Li~Yi, and Fisher Yu.
\newblock {ShapeNet: An Information-Rich 3D Model Repository}.
\newblock Technical Report arXiv:1512.03012 [cs.GR], 2015.

\bibitem{higgins_87}
H.~C. Longuet-Higgins.
\newblock Readings in computer vision: Issues, problems, principles, and
  paradigms.
\newblock chapter A Computer Algorithm for Reconstructing a Scene from Two
  Projections, pages 61--62. 1987.

\bibitem{wu_3dv13}
Changchang Wu.
\newblock Towards linear-time incremental structure from motion.
\newblock In {\em 3DV}, 2013.

\bibitem{blanz_siggraph99}
Volker Blanz and Thomas Vetter.
\newblock A morphable model for the synthesis of 3d faces.
\newblock In {\em SIGGRAPH}, 1999.

\bibitem{haene_3dv17}
Christian H{\"a}ne, Shubham Tulsiani, and Jitendra Malik.
\newblock Hierarchical surface prediction for 3d object reconstruction.
\newblock In {\em 3DV}, 2017.

\bibitem{lin_aaai18}
Chen-Hsuan Lin, Chen Kong, and Simon Lucey.
\newblock Learning efficient point cloud generation for dense 3d object
  reconstruction.
\newblock In {\em AAAI Conference on Artificial Intelligence ({AAAI})}, 2018.

\bibitem{wang_eccv18}
Nanyang Wang, Yinda Zhang, Zhuwen Li, Yanwei Fu, Wei Liu, and Yu-Gang Jiang.
\newblock Pixel2mesh: Generating 3d mesh models from single rgb images.
\newblock In {\em ECCV}, 2018.

\bibitem{groueix_cvpr18}
Thibault Groueix, Matthew Fisher, Vladimir~G. Kim, Bryan Russell, and Mathieu
  Aubry.
\newblock {AtlasNet: A Papier-M\^ach\'e Approach to Learning 3D Surface
  Generation}.
\newblock In {\em CVPR}, 2018.

\bibitem{lun_3dv17}
Zhaoliang Lun, Matheus Gadelha, Evangelos Kalogerakis, Subhransu Maji, and Rui
  Wang.
\newblock {3D} shape reconstruction from sketches via multi-view convolutional
  networks.
\newblock In {\em 3DV}, 2017.

\bibitem{tatarchenko_eccv16}
M.~Tatarchenko, A.~Dosovitskiy, and T.~Brox.
\newblock Multi-view {3D} models from single images with a convolutional
  network.
\newblock In {\em ECCV}, 2016.

\bibitem{richter_cvpr18}
Stephan Richter and Stephan Roth.
\newblock Matryoshka networks: Predicting 3d geometry via nested shape layers.
\newblock In {\em CVPR}, 2018.

\bibitem{rezende_nips16}
Danilo~Jimenez Rezende, S.~M.~Ali Eslami, Shakir Mohamed, Peter Battaglia, Max
  Jaderberg, and Nicolas Heess.
\newblock Unsupervised learning of 3d structure from images.
\newblock In {\em NIPS}, 2016.

\bibitem{gadelha_3dv17}
Matheus Gadelha, Subhransu Maji, and Rui Wang.
\newblock {3D} shape induction from {2D} views of multiple objects.
\newblock In {\em 3DV}, 2017.

\bibitem{jiajun_nips17}
Jiajun Wu, Yifan Wang, Tianfan Xue, Xingyuan Sun, William~T Freeman, and
  Joshua~B Tenenbaum.
\newblock {MarrNet: 3D Shape Reconstruction via 2.5D Sketches}.
\newblock In {\em NIPS}, 2017.

\bibitem{he2016deep}
Kaiming He, Xiangyu Zhang, Shaoqing Ren, and Jian Sun.
\newblock Deep residual learning for image recognition.
\newblock In {\em Proceedings of the IEEE conference on computer vision and
  pattern recognition}, 2016.

\bibitem{jaderberg2015spatial}
Max Jaderberg, Karen Simonyan, Andrew Zisserman, et~al.
\newblock Spatial transformer networks.
\newblock In {\em Advances in neural information processing systems}, 2015.

\bibitem{sun2018pix3d}
Xingyuan Sun, Jiajun Wu, Xiuming Zhang, Zhoutong Zhang, Chengkai Zhang, Tianfan
  Xue, Joshua~B Tenenbaum, and William~T Freeman.
\newblock Pix3d: Dataset and methods for single-image 3d shape modeling.
\newblock In {\em CVPR}, 2018.

\bibitem{miller2004graspit}
Andrew~T Miller and Peter~K Allen.
\newblock Graspit! a versatile simulator for robotic grasping.
\newblock {\em IEEE Robotics \& Automation Magazine}, 11(4), 2004.

\bibitem{gualtieri2016high}
Marcus Gualtieri, Andreas Ten~Pas, Kate Saenko, and Robert Platt.
\newblock High precision grasp pose detection in dense clutter.
\newblock In {\em IROS}, 2016.

\bibitem{chitta2012moveit}
Sachin Chitta, Ioan Sucan, and Steve Cousins.
\newblock Moveit!
\newblock {\em IEEE Robotics \& Automation Magazine}, 19, 2012.

\end{thebibliography}

\end{document}